\documentclass[10pt]{article}
\usepackage[utf8]{inputenc}
\usepackage[T1]{fontenc}
\usepackage{amsmath}
\usepackage{amsfonts}
\usepackage{amssymb}
\usepackage[version=4]{mhchem}
\usepackage{stmaryrd}

\title{Intensional Inheritance Between Concepts: \\ An Information-Theoretic Interpretation}

\author{Ben Goertzel}
\date{}

\begin{document}
\maketitle

\begin{abstract}
This paper addresses the problem of formalizing and quantifying the concept of ''intensional inheritance'' between two concepts.     We begin by conceiving the intensional inheritance of $W$ from $F$ as the amount of information the proposition "x is $F$ " provides about the proposition "x is $W$.   To flesh this out, we consider concepts $F$ and $W$ defined by sets of properties $\left\{F_{1}, F_{2}, \ldots, F_{n}\right\}$ and $\left\{W_{1}, W_{2}, \ldots, W_{m}\right\}$ with associated degrees $\left\{d_{1}, d_{2}, \ldots, d_{n}\right\}$ and $\left\{e_{1}, e_{2}, \ldots, e_{m}\right\}$, respectively, where the properties may overlap. We then derive formulas for the intensional inheritance using both Shannon information theory and algorithmic information theory, incorporating interaction information among properties.  We examine a special case where all properties are mutually exclusive and calculate the intensional inheritance in this case in both frameworks. We also derive expressions for $P(W \mid F)$ based on the mutual information formula.  Finally we consider the relationship between intensional inheritance and conventional set-theoretic "extensional" inheritance, concluding that in our information-theoretic framework, extensional inheritance emerges as a special case of intensional inheritance.
\end{abstract}

\section{Introduction}

The notion of "inheritance" between concepts is rich and multidimensional, with a long and diverse history, and no formalization is going to capture all the nuances.   Our goal here is to present a formalization of the notion of intensional inheritance that captures enough nuance in a coherent enough way to be useful for guiding reasoning in AI and AGI systems, with OpenCog Hyperon \cite{goertzel2023hyperon} and its Probabilistic Logic Networks \cite{goertzel2008PLN} reasoning system as the primary systems in mind.

\subsection{Intensional Inheritance}

In philosophy, broadly speaking, intension refers to the internal content or set of attributes that define a concept or term.  It is contrasted with extension, which refers to the set of instances that exemplify the concept \cite{fitting2006intensional}.  For example:

\begin{itemize}
\item The intension of "triangle" includes its defining properties, such as being a closed figure with three sides.
\item The extension of "triangle" includes all the actual triangles in existence.
\end{itemize}

\noindent Intension deals with the meaning or criteria of a concept, focusing on its descriptive or semantic aspects rather than its instances.

"Intensional inheritance" then refers to how concepts inherit or share defining properties or meanings in a hierarchical or structured way. This concept is commonly used in logic, linguistics, and ontological frameworks like FrameNet or SUMO.  For example:

\begin{itemize}
\item In a hierarchy where "dog" is a subclass of "mammal," the concept of "dog" inherits the intension of "mammal" (e.g., being warm-blooded, having fur, and giving live birth), while adding its own unique attributes (e.g., barking, wagging tails).
\item Similarly, "square" inherits the intension of "rectangle" (having four sides and right angles) but adds the property of all sides being equal.
\end{itemize}

\noindent Intensional inheritance allows for the structured organization of concepts where meanings are progressively specified while maintaining a connection to more general concepts.

\subsection{An Information-Theoretic Approach}

We propose to assess the degree of intensional inheritance between $W$ and $F$ by asking how much information the proposition "x is $F$ " provides about the proposition "x is $W$".

For instance if 

\begin{itemize}
\item  $W = \textrm{cat}$
\item $F = \textrm{animal}$ 
\end{itemize}

\noindent then we ask: 

\begin{itemize}
\item How much information does "x is 'animal'" give regarding "x is 'cat'"
\end{itemize}

More precisely: We consider two concepts $F$ and $W$, each defined by a set of properties:

\begin{itemize}
  \item Concept $F$ : Defined by properties $\left\{F_{1}, F_{2}, \ldots, F_{n}\right\}$ with degrees $\left\{d_{1}, d_{2}, \ldots, d_{n}\right\}$.
  \item Concept $W$ : Defined by properties $\left\{W_{1}, W_{2}, \ldots, W_{m}\right\}$ with degrees $\left\{e_{1}, e_{2}, \ldots, e_{m}\right\}$.
\end{itemize}

\noindent The degrees represent the probabilities or extents to which an element $x$ possesses each property. The properties $F_{i}$ and $W_{j}$ may overlap, introducing dependencies between $F$ and $W$.

We begin by deriving  a formula for the intensional inheritance of $W$ from $F$ using two separate but algebraically similar approaches:

\begin{itemize}
  \item Shannon information theory and associated interaction information.
  \item Algorithmic information theory and associated interaction information.
\end{itemize}

We also

\begin{itemize}
  \item Examine a special case where all properties $F_{i}$ and $W_{j}$ are mutually exclusive, and calculate the intensional inheritance in both frameworks.
  \item Derive expressions for the conditional probability $P(W \mid F)$ based on the mutual information formulas in both cases.
  \item Demonstrate that extensional inheritance (probabilistic subset relationships) emerge in this case as a special case of intensional inheritance when properties are singleton elements.
\end{itemize}

\section{Setup}

We begin with the following setup:

\begin{itemize}
  \item Concept $F$ : Defined by properties $\left\{F_{1}, F_{2}, \ldots, F_{n}\right\}$.
  \item Concept $W$ : Defined by properties $\left\{W_{1}, W_{2}, \ldots, W_{m}\right\}$.
  \item Degrees $d_{i}$ : The degree to which an element $x$ possesses property $F_{i}$.
  \item Degrees $e_{j}$ : The degree to which an element $x$ possesses property $W_{j}$.
  \item Overlap: Some properties may be common to both $F$ and $W$.
\end{itemize}

In this context, we will conceptualize the {\bf Intensional Inheritance} of $W$ from $F$ as, intuitively: The amount of information the proposition "x is $F$ " provides about the proposition "x is $W$ ".

This can be formalized in various ways depending on how one operationalizes the concept of "information."  We will explore two options here, using Shannon and algorithmic information.   One could probably unify these under a more general framework, considering a broader notion of an "information theory" as any theory satisfying a certain set of axioms.  The right way to do this seems to be to extend the use of Markov categories to model entropy \cite{perrone2023markov}, and broaden it to encompass algorithmic as well as statistical processes.  However, this would be another paper in itself, and for the present purposes we'll stick with these two well-understood options.

\section{Intensional Inheritance Using Shannon Information Theory}

\subsection{Preliminaries}

In Shannon information theory, the mutual information between $F$ and $W$ is defined as:

$$
I(F ; W)=H(W)-H(W \mid F)=H(F)+H(W)-H(F, W)
$$

\noindent where:

\begin{itemize}
  \item $H(F)$ : Entropy of $F$.
  \item $H(W)$ : Entropy of $W$.
  \item $H(F, W)$ : Joint entropy of $F$ and $W$.
  \item $H(W \mid F)$ : Conditional entropy of $W$ given $F$.
\end{itemize}

The interaction information, next, captures the dependencies among multiple variables beyond pairwise interactions \cite{VanDeCruys2011}. It adjusts the joint entropy to account for these dependencies.   For properties $F_{i}$ and $W_{j}$, the interaction information is included in the calculation of joint entropy $H(F, W)$. The total interaction information among properties can be expressed as:

$$
\text { Interaction Information }=\sum_{\text {all subsets }}(-1)^{|S|+1} I(S)
$$

\noindent where $I(S)$ is the mutual information among the properties in subset $S$.

\subsection{Derivation of Intensional Inheritance}

Given these quantities, we now work toward a derivation of a formula for intensional inheritance in terms of Shannon information, step by step.

We will derive a formula for $H(F, W)$ based on suitable assumptions, which then leads directly to a formula for  $P(F, W)$, which is our conceptualization of intensional inheritance and our goal.

Assuming independence among $F_{i}$ (which we'll relax later), the entropy of $F$ is:

$$
H(F)=-\sum_{f} P(F=f) \log P(F=f)
$$

\noindent But $F$ is defined by its properties $F_{i}$. If the properties are independent, then:

$$
P(F)=\prod_{i=1}^{n} P\left(F_{i}\right)
$$

\noindent Similarly, the entropy $H(F)$ can be calculated based on the individual entropies and interaction information.

The joint entropy $H(F, W)$ accounts for the entropy of both $F$ and $W$, including their dependencies:

$$
H(F, W)=H(F)+H(W)-I(F ; W)
$$

But since $I(F ; W)$ depends on the interaction among properties, we need to incorporate interaction information.

The mutual information $I(F ; W)$ including interaction information among properties is:

$$
I(F ; W)=\left(\sum_{i} H\left(F_{i}\right)+\sum_{j} H\left(W_{j}\right)-H(F, W)\right)-\text { Interaction Information }
$$

\noindent where:

\begin{itemize}
  \item $\sum_{i} H\left(F_{i}\right)$ : Sum of entropies of individual properties $F_{i}$.
  \item $\sum_{j} H\left(W_{j}\right)$ : Sum of entropies of individual properties $W_{j}$.
  \item $H(F, W)$ : Joint entropy of all properties, accounting for their dependencies.
  \item Interaction Information: Adjusts for the dependencies among properties.
\end{itemize}

To compute $I(F ; W)$, we will proceed as follows.   First, we calculate individual entropies $H\left(F_{i}\right)$ and $H\left(W_{j}\right)$.   For each property $F_{i}$ and $W_{j}$ :

$$
\begin{gathered}
H\left(F_{i}\right)=-d_{i} \log d_{i}-\left(1-d_{i}\right) \log \left(1-d_{i}\right) \\
H\left(W_{j}\right)=-e_{j} \log e_{j}-\left(1-e_{j}\right) \log \left(1-e_{j}\right)
\end{gathered}
$$

Then, we calculate joint entropy $H(F, W)$.   The joint entropy involves the probabilities of all combinations of $F_{i}$ and $W_{j}$, adjusted for dependencies.  Interaction information is calculated based on the dependencies among properties. For instance, if certain properties are dependent or overlap, their joint probabilities differ from the product of their marginals.  To compute mutual information $I(F ; W)$, we substitute the calculated values into the mutual information formula, accounting for interaction information.

In detail:

$$
I(F ; W)=H(W)-H(W \mid F)
$$

Rearranged:

$$
H(W \mid F)=H(W)-I(F ; W)
$$

The conditional entropy $H(W \mid F)$ is:

$$
H(W \mid F)=-\sum_{f} P(F=f) \sum_{w} P(W=w \mid F=f) \log P(W=w \mid F=f)
$$

Assuming uniform distribution (for simplification), this works out as follows.

If $W$ has $k$ possible values and is uniformly distributed, $H(W)=\log k$, then:

$$
H(W \mid F)=\log k-I(F ; W)
$$

\noindent and the average conditional probability is:

$$
\bar{P}(W \mid F)=2^{-H(W \mid F)}=2^{-(\log k-I(F ; W))}=k^{-1} \cdot 2^{I(F ; W)}
$$

\noindent yielding the final formula:

$$
P(W \mid F)=P(W) \cdot 2^{I(F ; W)}
$$

Of course, if there is prior knowledge violating the simplifying assumption of a uniform distribution, it should be deployed and one will obtain a different result.

\section{Intensional Inheritance Using Algorithmic Information Theory}

An analogous derivation may be given using algorithmic information theory \cite{li2019algorithmic}

In algorithmic information theory, the mutual information between $F$ and $W$ is defined as:

$$
I(F: W)=K(W)-K(W \mid F)
$$

\noindent where:

\begin{itemize}
  \item $\quad K(W)$ : Kolmogorov complexity of $W$.
  \item $\quad K(W \mid F)$ : Conditional Kolmogorov complexity of $W$ given $F$.
\end{itemize}

The interaction information among properties may be incorporated as follows:

$$
\begin{aligned}
K(W) & =\sum_{j=1}^{m} K\left(W_{j}\right)-\operatorname{Inter}_{W} \\
K(F) & =\sum_{i=1}^{n} K\left(F_{i}\right)-\text { Inter }_{F} \\
K(F, W) & =\sum_{k=1}^{n+m} K\left(P_{k}\right)-\text { Inter }_{F, W}
\end{aligned}
$$

\noindent where:

\begin{itemize}
  \item $K\left(P_{k}\right)$ : Kolmogorov complexities of all properties.
  \item $\operatorname{Inter}_{F}, \operatorname{Inter}_{W}, \operatorname{Inter}_{F, W}$ : Interaction information among properties.
\end{itemize}

We may compute mutual information via $I(F: W)$ :

$$
I(F: W)=K(W)-K(W \mid F)
$$

But since:

$$
K(W \mid F)=K(F, W)-K(F)
$$

\noindent substituting we obtain:

$$
I(F: W)=K(W)-[K(F, W)-K(F)]=K(W)+K(F)-K(F, W)
$$

\noindent and considering the interaction terms we obtain:

$$
I(F: W)=\left(\sum_{i=1}^{n} K\left(F_{i}\right)+\sum_{j=1}^{m} K\left(W_{j}\right)-\sum_{k=1}^{n+m} K\left(P_{k}\right)\right)+\left(\operatorname{Inter}_{F, W}-\operatorname{Inter}_{F}-\operatorname{Inter}_{W}\right)
$$

Using the relationship between algorithmic mutual information and conditional probability, we then calculate the algorithmic probability of $W$ is:

$$
P(W) \approx 2^{-K(W)}
$$

\noindent and similarly, the conditional probability:

$$
P(W \mid F) \approx 2^{-K(W \mid F)}
$$

We can express $K(W \mid F)$ in terms of Mutual Information:

$$
K(W \mid F)=K(W)-I(F: W)
$$

\noindent and substitute back into $P(W \mid F)$ :

$$
P(W \mid F) \approx 2^{-K(W)+I(F: W)}=P(W) \cdot 2^{I(F: W)}
$$

\noindent obtaining the final formula:

$$
P(W \mid F) \propto P(W) \cdot 2^{I(F: W)}
$$

\noindent which of course is formally very similar to the Shannon information case.

\section{Special Case: Mutually Exclusive Properties}

We now consider a special case where all properties $F_{i}$ and $W_{j}$ are mutually exclusive.   This is a simple enough situation that we can get an explicit elementary formula.

\subsection{Assumptions}

Mutual Exclusivity:
\begin{itemize}
  \item $F_{i} \cap F_{j}=\emptyset$ for $i \neq j$.
  \item $W_{i} \cap W_{j}=\emptyset$ for $i \neq j$.
  \end{itemize}

\noindent Degrees:

\begin{itemize}
  \item Each property has degree $p=\frac{1}{s}$.
  \item $s$ is the total number of unique properties.
\end{itemize}

\noindent Overlap:
\begin{itemize}
  \item There are $k$ properties common to both $F$ and $W$.
\end{itemize}

\subsection{Calculations Using Shannon Information Theory}

\paragraph{Total Unique Properties:}

$$
s=n+m-k
$$

\paragraph{Degree of Each Property:}

$$
p=\frac{1}{s}
$$

\paragraph{Probability of $F$ :}

$$
P(F)=n \cdot p=\frac{n}{s}
$$

\paragraph{Probability of $W$ :}

$$
P(W)=m \cdot p=\frac{m}{s}
$$

\paragraph{Probability of $F \cap W$ :}

$$
P(F \cap W)=k \cdot p=\frac{k}{s}
$$

\paragraph{Conditional Probability}
$$
P(W \mid F)=\frac{P(F \cap W)}{P(F)}=\frac{\frac{k}{s}}{\frac{n}{s}}=\frac{k}{n}
$$

\subsection{Calculations Using Algorithmic Information Theory}

Assuming equal complexities for properties:

\begin{itemize}
  \item $K\left(F_{i}\right)=\log s$
  \item $K\left(W_{j}\right)=\log s$
  \end{itemize}
  
\noindent and total complexities:
  
  \begin{itemize}
  \item $K(F)=\log n$
  \item $K(W)=\log m$
  \item $K(F \cap W)=\log k$
\end{itemize}

\noindent we then have

\paragraph{Mutual Information}
$$
I(F: W)=K(W)-K(W \mid F)=\log m-\left(\log m-\log \left(\frac{k}{n}\right)\right)=\log \left(\frac{k}{n}\right)
$$

But since $P(W \mid F)=\frac{k}{n}$, we have:

$$
I(F: W)=\log P(W \mid F)
$$

\paragraph{Conditional Probability}

$$
P(W \mid F)=P(W) \cdot 2^{I(F: W)}=\frac{m}{s} \cdot \frac{k}{n}
$$

\section{Extensional Inheritance as a Special Case}

Finally, we note the relation between intensional inheritance as we have formulated it and simple "extensional inheritance" in the sense of overlapping set membership.   It  is immediately obvious that our notion of intensional inheritance is broad enough to include extensional inheritance as a special case, which is convenient in terms of formal and conceptual analysis and software implementation, though different from how things have been done in commonsense reasoning systems like PLN \cite{goertzel2008PLN} and NARS \cite{wang2006rigid} in the past.

That is, it is immediate to observe that: When properties are singleton elements (e.g., $F_{i}=\left\{x_{i}\right\}$ ), the intensional inheritance as defined here reduces to extensional inheritance.

The conceptual relationship as envisioned here then looks like:

\begin{itemize}
  \item Extensional Inheritance: A probabilistic subset relationship where knowing $x$ is in $F$ directly informs us about $x$ being in $W$.
  \item Intensional Inheritance: Generalizes this by considering overlapping properties and degrees.
\end{itemize}

\noindent Quite simply: In the case where each property corresponds to a unique element, and degrees are either 0 or 1 , the intensional inheritance formula simplifies to the extensional case.

\section{Conclusion}
We have derived detailed formulas for the intensional inheritance of $W$ from $F$ using both Shannon information theory and algorithmic information theory, incorporating interaction information among properties. In the special case of mutually exclusive properties, the formulas simplify considerably.  Finally, we observe that intensional inheritance encompasses extensional inheritance as a special case. 

This framework provides a quantitative method to assess how knowledge of one concept influences our understanding of another, accounting for the complex interplay of properties and their degrees.

\section{Acknowledgements} The author would like to thank Nil Geisweiller for posing the problem of reducing intensional and extensional inheritance elegantly to a single thing, at the December 2025 Hyperon workshop in Florianopolis, which is what spurred the ideas presented here.   And also would like to thank Pei Wang for, back in the late 1990s, spurring him to start thinking about the relationship between intensional and extensional inheritance in the first place ... although Pei's formal view of intension differs a fair bit from the one presented here.

\bibliographystyle{alpha}
\bibliography{intension}

\end{document}